\documentclass[onecolumn]{article}
\usepackage{lmodern}
\usepackage[list=true]{subcaption}
\usepackage{tikz}
\usetikzlibrary{decorations.pathreplacing}
\usepackage{xcolor}
\usepackage{tensor}
\usepackage{tabu}
\usepackage{listings}
\usepackage[linesnumbered,noend]{algorithm2e}
\usepackage{cancel}
\usepackage[section]{placeins}
\usepackage{amssymb,amsmath}
\usepackage{mathtools}
\usepackage{ifxetex,ifluatex}
\usepackage{fixltx2e} 
\ifnum 0\ifxetex 1\fi\ifluatex 1\fi=0 
  \usepackage[T1]{fontenc}
  \usepackage[utf8]{inputenc}
\else 
  \ifxetex
    \usepackage{mathspec}
    \usepackage{xltxtra,xunicode}
  \else
    \usepackage{fontspec}
  \fi
  \defaultfontfeatures{Mapping=tex-text,Scale=MatchLowercase}
  
\usepackage{fancyhdr}
\fi
\IfFileExists{upquote.sty}{\usepackage{upquote}}{}
\IfFileExists{microtype.sty}{%
\usepackage{microtype}
\UseMicrotypeSet[protrusion]{basicmath} 
}{}
\ifxetex
  \usepackage[setpagesize=false, 
              unicode=false, 
              xetex]{hyperref}
\else
  \usepackage[unicode=true]{hyperref}
\fi
\hypersetup{breaklinks=true,
            bookmarks=true,
            pdfauthor={Jad Nohra},
            pdftitle={Uniqueness of Minima of a Certain Least Squares Problem},
            colorlinks=true,
            citecolor=blue,
            urlcolor=blue,
            linkcolor=magenta,
            pdfborder={0 0 0}}
\title{Uniqueness of Minima of a Certain Least Squares Problem}
\author{  \small Jad Nohra}
\date{}

\begin{document}
\maketitle

\begin{abstract}
  This paper is essentially an exercise in studying the
  minima of a certain least squares optimization using the second partial derivative test. The motivation is to
  gain insight into an optimization-based solution to the problem of tracking human
  limbs using IMU sensors.
\end{abstract}

\section{Introduction}\label{introduction}
We study\footnote{As suggested to us by Teodor Cioac\u a and Horea
  C\u ar\u amizaru, whom we both thank.} the minima of a specific least squares problem using the second partial derivative test.
The problem's origin is an optimization-based solution proposed in (Seel, Schauer, and Raisch 2012) to enable robust tracking of human
limbs using IMU sensors. The original problem works with 6 dof rigid-body limbs in three
dimensional space, but we shall instead work on a planar version of it to simplify the
analysis. This is harmless given that our purpose is to study the uniqueness of
minima: if the minima are not unique in the planar case
problem, they are also not so for the spatial one.

\section{Analysis}\label{analysis}

\subsection{Problem Statement and
Notation}\label{problem-statement-and-notation}

We consider the nonlinear optimization problem of minimizing the sum of
square errors objective function \(O_n(\theta_1,\theta_2)\), where \(n\)
is the number of samples and \(\theta_i\) are angles. Each sample \(s\)
is a six dimensional vector that determines the sample's error and is
denoted by\footnote{We shall freely drop the sample index when
  convenient.}
\((\tensor[^s]{w}{_{11}},\tensor[^s]{w}{_{12}},\tensor[^s]{w}{_{13}},\tensor[^s]{w}{_{21}},\tensor[^s]{w}{_{22}},\tensor[^s]{w}{_{23}})\).
Given the above, \(O_n\) is given by

\begin{align}
 O_n(\theta_1,\theta_2) &= \displaystyle \sum_{s=1}^{n} (\tensor[^s]{d}{}(\theta_1,\theta_2))^2, \\
 \tensor[^s]{d}{}(\theta_1,\theta_2) &= p(\theta_1, \tensor[^s]{w}{_{1}}) - p(\theta_2, \tensor[^s]{w}{_{2}}), \\
 p(\theta_i, \tensor[^s]{w}{_{i}}) &= [\tensor[^s]{w}{_{i1}} \sin(\theta_i) - \tensor[^s]{w}{_{i3}} \cos(\theta_i)]^2 + (\tensor[^s]{w}{_{i2}})^2.
\end{align}

This corresponds to equation (1) in (Seel, Schauer, and Raisch 2012) up
to variable names after the switching to a two-dimensional planar hinge
problem where only a single angle has to be determined, moving to
spherical coordinates with

\begin{equation}
j_i = \left( \begin{array}{c} \cos(\theta_i) \\ 0 \\ \sin(\theta_i) \end{array} \right),
\end{equation}

and application of trigonometric identities.

We shall mostly skip the dependent variables for functions within
expressions and move any indices that need to be retained under the
letter. Per example we shall write \(p_1\) to mean \(p(\theta_1, w_1)\).
Additionally, we shall employ Newton's dot even for a partial derivative
when there is no ambiguity.

We now procede to finding the stationary points and characterizing them
using the second partial derivative test.

\subsection{Insight by Computer}\label{insight-by-computer}

Despite the simplicity of the problem, a direct attempt using symbolic
mathematics software is undermined by the fact that the symbolic
expressions generated for the relevant quantities for even the
single-sample problem \(O_1(\theta_1, \theta_2)\) are unworkable for a
human as one can judge from their length and form in the appendix.

\subsection{Analytical Solution of the Single-Sample
Problem}\label{analytical-solution-of-the-single-sample-problem}

\subsubsection{Preliminaries}\label{preliminaries}

It is a fact that the sum of two equal period sinusoids is another
sinusoid with the same period. Such sinusoids are known in Physics as
\(\textit{phasors}\), and the fact can be proved using trigonometric
identities and is expressed by:

\begin{equation}
A \sin(x) - B \cos(x) = \sqrt{A^2+B^2} \cos(x + \text{atan}[{\frac{A}{B}}]).
\end{equation}

Given this, let us define a number of functions that will prove helpful
when applying the chain rule during differentiation:

\begin{align}
r(\theta_i, w_i) &= (\theta_i + \text{atan}[\frac{w_{i1}}{w_{i3}}]). \\
s(w_i) &= (w_{i1}^2 + w_{i3}^2). \\
t(\theta_i, w_i) &= w_{i1} \sin(\theta_i) - w_{i3} \cos(\theta_i), \\
 &= \sqrt{s(w_i)} \cos(r(\theta_i, w_i)).\\
p(\theta_i, w_i) &= t^2(\theta_i, w_i) + (w_{i2})^2.
\end{align}

Due to the abundance of \(\sin(k\,r_i)\) and \(\cos(k\,r_i)\) where
\(k\) is an integer in what follows, we shall shorten such terms in this
manner:

\begin{align}
\cos(k\,r_i) &\twoheadrightarrow \left[{k \atop\text{co}_i}\right],\nonumber \\
\cos(k_1\,r_i)\,\cos(k_2r_j) &\twoheadrightarrow \left[{k_1 \atop\text{co}_i}{k_2 \atop\text{co}_j}\right],\nonumber
\end{align}

and similarly for the sine function. In this notation, the known
half-angle trigonometric identities are

\begin{align}
 \left[{k \atop\text{co}_i}{k \atop\text{si}_i}\right] &= \frac{1}{2} \left[{2k \atop\text{si}_i}\right].\nonumber\\
 \left[{k \atop\text{co}_i}\right]^2 &= \frac{1}{2} + \frac{1}{2} \left[{2k \atop\text{co}_i}\right].\nonumber
\end{align}

\subsubsection{Partial Derivatives and Hessian
Determinant}\label{partial-derivatives-and-hessian-determinant}

\paragraph{First Partial Derivatives}\label{first-partial-derivatives}

\begin{align}
\frac{\partial r_i}{\partial \theta_i} &= 1. \nonumber\\
\frac{\partial t_i}{\partial \theta_i} &= \sqrt{s_i}\,\dot{\cos}(r_i)\nonumber\\
 &= \sqrt{s_i}\,[-\sin(r_i)]\,\dot{r_i}\nonumber\\
 &= (-\sqrt{s_i})\,\sin(r_i). \nonumber\\
\frac{\partial p_i}{\partial \theta_i} &= 2\,t_i\,\dot{t_i}\nonumber \\
 &= 2\,[\sqrt{s_i}\,\cos(r_i)]\,[(-\sqrt{s_i})\,\sin(r_i)]\nonumber \\
 &= (-s_1)\,\sin(2r_i). \nonumber\\
\frac{\partial d}{\partial \theta_1} &= \frac{\partial (p_1-p_2)}{\partial \theta_1}\nonumber \nonumber\\
 &= (-s_1)\,\sin(2r_1). \\
\frac{\partial d}{\partial \theta_2} &= \frac{\partial (p_1-p_2)}{\partial \theta_2}\nonumber \nonumber\\
 &= (s_2)\,\sin(2r_2).\nonumber \\
\frac{\partial O_1}{\partial \theta_1} &= 2\,d\,\frac{\partial d}{\partial \theta_1}\nonumber \\
 &= -2s_1\,d\,\sin(2r_1). \\
 \frac{\partial O_1}{\partial \theta_2} &= 2\,d\,\frac{\partial d}{\partial \theta_2}\nonumber \\
  &= 2s_2\,d\,\sin(2r_2).
\end{align}

We are now in a position to write the Jacobian as

\begin{equation}
 \text{Jac }{O_1} = \left(-2s_1\,d\left[{2\atop\text{si}_1}\right],\,\, 2s_2\,d\left[{2\atop\text{si}_2}\right] \right).
\end{equation}

Let us additionally expand \(\dfrac{\partial O}{\partial \theta_i}\) in
terms of trigonometric functions for later use.

\begin{align}
d &= (t_1^2+w_{12}^2) - (t_2^2+w_{22}^2)\nonumber \\
 &= \frac{s_1}{2} + \frac{s_1}{2} \left[2\atop\text{co}_1\right] + w_{12}^2 - \frac{s_2}{2} - \frac{s_2}{2} \left[2\atop\text{co}_2\right] - w_{22}^2\nonumber \\
 &= \frac{1}{2} \left( c + s_1\left[2\atop\text{co}_1\right] - s_2\left[2\atop\text{co}_2\right] \right).
\end{align}

where

\begin{equation}
 c(w_1,w_2) = [ s_1+2w_{12}^2-s_2-2w_{22}^2 ].
\end{equation}

Hence,

\begin{align}
\frac{\partial O_1}{\partial \theta_1}
 &= (-2\,s_1)\left[\frac{1}{2} \left( c + s_1\left[2\atop\text{co}_1\right] - s_2\left[2\atop\text{co}_2\right] \right)\right]\left[2\atop\text{si}_1\right]\nonumber \\
 &= (-s_1)\left[2\atop\text{si}_1\right]\left( c + s_1\left[2\atop\text{co}_1\right] - s_2\left[2\atop\text{co}_2\right]\right), \label{eq:note39}\\
 &=  (-s_1)\left( c\left[2\atop\text{si}_1\right] + \frac{1}{2}s_1\left[{4\atop\text{si}_1}\right] - s_2\left[{2\atop\text{si}_1}{2\atop\text{co}_2}\right]\right).
\end{align}

Similarly, we have for \(\dfrac{\partial O_1}{\partial \theta_2}\):

\begin{align}
\frac{\partial O_1}{\partial \theta_2}
 &= (s_2)\left[2\atop\text{si}_2\right]\left( c + s_1\left[2\atop\text{co}_1\right] - s_2\left[2\atop\text{co}_2\right]\right),\\
 &=  (s_2)\left( c\left[2\atop\text{si}_2\right] +
 s_1\left[{2\atop\text{co}_1}{2\atop\text{si}_2}\right] - \frac{1}{2}s_2\left[{4\atop\text{si}_2}\right] \right).
\end{align}

\paragraph{Second Partial Derivatives}\label{second-partial-derivatives}

Making use of the various derivations from the last section we have:

\begin{align}
\frac{\partial^2 O_1}{\partial \theta_1^2}
 &= \frac{\partial \left(-2s_1\,d\,\sin(2r_1)\right)}{\partial \theta_1}, \nonumber \\
 &= -2s_1 \frac{\partial(d\,\sin(2r_1))}{\partial \theta_1}, \nonumber \\
 &= -2s_1 \left[ d(2\cos(2r1) + (-s_1\sin(2r_1)(\sin(2r_1) \right], \nonumber \\
 &= (-2s_1) \left(2d\left[2\atop\text{co}_1\right] -s_1\left[2\atop\text{si}_1\right]^2\right), \nonumber \\
  &= (-2s_1) \left(2d\left[2\atop\text{co}_1\right] -s_1\left(1-\frac{1}{2}-\frac{1}{2}\left[4\atop\text{co}_1\right]\right)\right), \nonumber \\
 &=  (-2s_1) \left( \frac{-s_1}{2} + 2d\left[2\atop\text{co}_1\right] + \frac{s_1}{2}\left[4\atop\text{co}_1\right]\right).
\end{align}

A full expansion of the expression along with the half-angle identity
gives the alternative form

\begin{equation}
\frac{\partial^2 O_1}{\partial \theta_1^2}
 = (-2s_1) \left( c\left[2\atop\text{co}_1\right] + s_1\left[4\atop\text{co}_1\right] - s_2\left[{2\atop\text{co}_1}{2\atop\text{co}_2}\right] \right).
\end{equation}

In a similar vein we obtain the derivative relative to \(\theta_2\):

\begin{align}
\frac{\partial^2 O_1}{\partial \theta_2^2} = (2s_2) \left( \frac{s_2}{2} + 2d\left[2\atop\text{co}_2\right] - \frac{s_2}{2}\left[4\atop\text{co}_2\right]\right), \\
= (2s_2) \left( c\left[2\atop\text{co}_2\right] - s_2\left[4\atop\text{co}_2\right] + s_1\left[{2\atop\text{co}_1}{2\atop\text{co}_2}\right] \right).
\end{align}

It is obvious that the order of differentiation does not matter when
considering the partial derivative with respect to
\(\partial\theta_1\partial\theta_2\). This derivative is given by

\begin{align}
\frac{\partial^2 O_1}{\partial\theta_2\partial\theta_1}
 &= \frac{\partial \left( -2s_1\,d\sin(2r_1)  \right)}{\partial\theta_2}, \nonumber \\
 &= \left( -2s_1\sin(2r_1) \right) \frac{\partial d}{\partial\theta_2}, \nonumber \\
 &= -2{s_1}{s_2} \left[{2\atop\text{si}_1}{2\atop\text{si}_2}\right].
\end{align}

\newcommand{\mdh}[1]{ \lvert{\tensor*[_ {#1}]{H}{}}\rvert }
\newcommand{\mdhex}[2]{ \tensor[^{#2}]{\lvert{\tensor*[_ {#1}]{H}{}}}{} \rvert }

Finally, we denote the problem's Hessian's determinant by

\begin{align}
\lvert{H}\rvert
 &= \underbrace{ \frac{\partial^2 O_1}{\partial \theta_1^2}\frac{\partial^2 O_1}{\partial \theta_2^2} }_{\mdh{L}} - \underbrace{ {\left[\frac{\partial^2 O_1}{\partial \theta_1\partial \theta_2}\right]}^2 }_{\mdh{R}}, \\
 &= \mdh{L} - 4 \left( s_1 \left[2\atop\text{si}_1\right] \right)^2 \left( s_2 \left[2\atop\text{si}_2\right] \right)^2.
\end{align}

\subsubsection{Stationary Points}\label{stationary-points}

\paragraph{Zeros of First Partials}\label{zeros-of-first-partials}

The expressions of the first partials are the product of three terms,
the values of \(\theta_i\) at which we obtain zeros are the union of
three sets. The first terms \((-s_1)\) and \((s_2)\) are not a function
of \(\theta_i\) so we consider the case \((s_1 = s_2 = 0)\) as
degenerate going forward and ignore it; since \(s_i\) is non-negative by
definition, we assume from now on that

\begin{equation}
s_i > 0.
\end{equation}

\newcommand{\mzt}[2]{ \tensor*[^{\theta_{#1}}_{#2}]{Z}{} }
\newcommand{\mso}[2]{ \tensor*[^{O_{#1}}_{#2}]{S}{} }
\newcommand{\msos}[2]{ \tensor*[^{}_{#1}]{S}{} }

This leaves us with two sets per partial. Paying close attention to
similarities and differences between the two partials, we denote the
sets as

\begin{align}
  \mzt{i}{1} &: \{\theta_i : \left[2\atop\text{si}_i\right] = 0 \}, \\
  \mzt{}{2} &: \{(\theta_1,\theta_2) : \left( c + s_1\left[2\atop\text{co}_1\right] - s_2\left[2\atop\text{co}_2\right]\right) = 0 \}.
\end{align}

As we shall see, the sets are more tersely described if we focus on
\(2r_i\) instead of on \(\theta_i\) and this is harmless since the
former are merely offset and scaled functions of the latter. Out of
terseness as well we let

\begin{align}
\alpha_i(w_i) &= \text{atan}[\frac{w_{i1}}{w_{i3}}], \\
\beta(w) &= 2w_{12}^2-2w_{22}^2.
\end{align}

\(\mzt{i}{1}\) has the simple solution

\begin{align}
\sin(2r_i) &= 0, \\
2r_i &= k\pi,\ (k \in \mathbb{Z}).
\end{align}

\paragraph{First Stationary Set}\label{first-stationary-set}

The intersection of \(\mzt{i}{1}\) directly gives us a first set of
stationary points in the \((\theta_1,\theta_2)\) plane as

\begin{align}
 \msos{1}{1}
  &: \{ (  - \alpha_1 + [\frac{k_1\pi}{2}], - \alpha_2 + [\frac{k_2\pi}{2}] ) : k_1,k_2 \in \mathbb{Z} \}.
\end{align}

Let us conduct the second partial derivative test on this set. We first
find the cases to consider. Since \(2r_i = k\pi\), all sines involved
are zero. On the other hand the cosines fall in the set \(\{-1,1\}\) and
we need to consider both cases. The first one is

\begin{align*}
  \cos(2r_i) &= -1, \\
  2r_i = 2k_i\pi+\pi, \\
  r_i = k_i\pi+\frac{\pi}{2}.
\end{align*}

The second is similar and we find that we need to consider the even and
odd values of \(k_i\) separately. This gives a total combination of four
cases as follows:

\[
\begin{dcases}
  k_1 \text{ even }, k_2 \text{ even}, \\
  k_1 \text{ odd }, k_2 \text{ even}, \\
  k_1 \text{ even }, k_2 \text{ odd}, \\
  k_1 \text{ odd }, k_2 \text{ odd}.
\end{dcases}
\]

Treating all four cases simultaneously with the aid of an unorthodox but
straightforward notation, we have that \(\mdhex{L}{\msos{1}{1}}\) equals
to\footnote{By \(\mdhex{L}{\msos{1}{1}}\) We mean \(\mdhex{L}{}\)
  evaluated at points in \(\msos{1}{1}\).}

\newcommand{\mcc}[4]{\kern-0.5ex\left[\kern-1.5ex \begin{array}{c}\scriptscriptstyle{#1}\\\scriptscriptstyle{#2}\\\scriptscriptstyle{#3}\\\scriptscriptstyle{#4}\end{array} \kern-1.5ex\right]\kern-0.5ex }

\begin{align*}
 & -s_1 ( 2c\mcc{+1}{-1}{+1}{-1} + 2s_1 \mcc{1}{1}{1}{1} -2s_2 \mcc{+1}{-1}{+1}{-1} \mcc{+1}{+1}{-1}{-1} )
  s_2( 2c \mcc{+1}{+1}{-1}{-1} - 2s_2 \mcc{1}{1}{1}{1} +2s_1 \mcc{+1}{-1}{+1}{-1} \mcc{+1}{+1}{-1}{-1} ) \\
&=
(-s_1 s_2) ( 2s_1 + 2c \mcc{+1}{-1}{+1}{-1} -2s_2 \mcc{+1}{-1}{-1}{+1} ) ( -2s_2 + 2c \mcc{+1}{+1}{-1}{-1} + 2s_1 \mcc{+1}{-1}{-1}{+1} )
\end{align*}

Expanding \(c\) we get after a few steps that

\begin{align}
\mdhex{L}{\msos{1}{1}}
 &= (-s_1 s_2) (\mcc{4s_1}{0}{4s_1}{0} + \mcc{-4s_2}{4s_2}{0}{0} + \mcc{2\beta}{-2\beta}{2\beta}{-2\beta}) (\mcc{4s_1}{0}{-4s_1}{0} + \mcc{-4s_2}{-4s_2}{0}{0} + \mcc{2\beta}{2\beta}{-2\beta}{-2\beta}), \nonumber \\
  &= \begin{drcases} (- s_1 s_2) \mcc{+1}{-1}{-1}{+1} \mcc{(4s_1-4s_2+2\beta)^2}{(4s_2-2\beta)^2}{(4s_1+2\beta)^2}{(-2\beta)^2} \quad \end{drcases} \begin{array}{c}{< 0}\\{> 0}\\{> 0}\\{< 0}\end{array}.
\end{align}

Since it is clear that \(\mdhex{R}{\msos{1}{1}}\) amounts to zero
unconditionally, we reach the result that \(\mdhex{}{\msos{1}{1}}\) does
not depend on the sample's data. Ignoring degenerate cases, we see that
we have saddle points for the first and fourth cases, and extrema for
the second and third.

We continue the test and determine the nature of the extrema by
examining the sign of \(\frac{\partial^2 O_1}{\partial \theta_1^2}\),
the expression of which we already worked out during the previous
calculation and which is

\begin{equation}
\tensor[^{\msos{1}{1}}]{\frac{\partial^2 O_1}{\partial \theta_1^2} }{} =
 (- s_1) \mcc{4s_1-4s_2+2\beta}{4s_2-2\beta}{4s_1+2\beta}{-2\beta}.
\end{equation}

Let us derive for the extremal cases the conditions under which we
obtain maxima, which is the relevant case as we shall see later. For the
second case we require:

\begin{align*}
(-s_1) (4s_2-2 [2w_{12}^2 - 2w_{22}^2] ) < 0, \\
 w_{21}^2+w_{23}^2 - w_{12}^2 + w_{22}^2 > 0, \\
 w_{12}^2 < w_{21}^2+w_{23}^2+w_{22}^2.
\end{align*}

We obtain a similar result for the third case and we conclude that the
extrema are maxima under the elegant condition

\begin{equation}
w_{12}^2 < \|w_2\|^2 \text{ and } w_{21}^2 < \|w_1\|^2. \label{eqn:note49}
\end{equation}

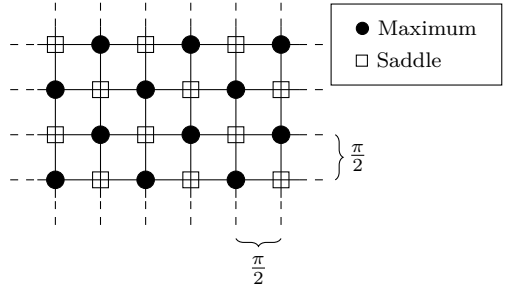
\begin{figure}
\centering
\begin{tikzpicture}[scale=0.6]
 \def\sc{0.7};
 \def\sr{0.9};
  \foreach \i in {0,...,3}
  {
    \draw[thin,dashed] (-1,\i) -- (-0.4,\i); \draw[thin] (-0.4,\i) -- (5.4,\i); \draw[thin,dashed] (5.4,\i) -- (6,\i);
  }
  \foreach \i in {0,...,5}
  {
    \draw[thin,dashed] (\i,-1) -- (\i,-0.4); \draw[] (\i,-0.4) -- (\i,3.4); \draw[thin,dashed] (\i,3.4) -- (\i,4);
  }

  \foreach \y in {0,2}
  {
    \foreach \x in {0,2,4}
    {
        \node[circle,fill,draw,scale=\sc,thin] at (\x,\y) {};
    }
    \foreach \x in {1,3,5}
    {
        \node[rectangle,draw,scale=\sr,thin] at (\x,\y) {};
    }
  }
  \draw [decorate,decoration={brace,amplitude=3pt},rotate around={180:(6.2,0.5)},=180] (6.2,0) -- (6.2,1);
  \node at (6.7,0.5) {$\frac{\pi}{2}$};

  \draw [decorate,decoration={brace,amplitude=3pt},rotate around={180:(4.5,-1.3)},=180] (4,-1.3) -- (5,-1.3);
  \node at (4.5,-2.0) {$\frac{\pi}{2}$};

  \foreach \y in {1,3}
  {
    \foreach \x in {1,3,5}
    {
        \node[circle,fill,draw,scale=\sc,thin] at (\x,\y) {};
    }
    \foreach \x in {0,2,4}
    {
        \node[rectangle,draw,scale=\sr,thin] at (\x,\y) {};
    }
  }

  \node[draw] at (8,3) {
    \begin{tabular}{l} \begin{tikzpicture}\node[circle,fill,draw,scale=\sc*0.7,thin] at (0,0) {};\end{tikzpicture} \smaller{Maximum} \\ \begin{tikzpicture}\node[rectangle,draw,scale=\sr*0.7,thin] at (0,0) {};\end{tikzpicture} \smaller{Saddle} \end{tabular}};
\end{tikzpicture}
\caption{ The grid of stationary points $\mso{1}{1}$.} \label{fig:note50}
\end{figure}

Pictorially, we obtain a simple grid of saddle points and maxima for
\(\msos{1}{1}\) as shown in figure \ref{fig:note50}.

\paragraph{Second Stationary Set}\label{second-stationary-set}

Unlike the discrete set above, the second set of zeros already
implicitly links the two variables into a curve which we shall study. By
this we have that \(\msos{2}{1}\) is identical to \(\mzt{}{2}\) and is
characterized by

\begin{equation}
\cos(2r_1) = \frac{-c}{s_1} + \frac{s_2}{s_1}\cos(2r_2). \label{eq:np55}
\end{equation}

Let us first study the set's existence, which is obviously determined by
the condition that the right-hand-side of the equation above is within
\([-1,1]\).

We have that

\begin{align*}
\frac{-c}{s_1}
 &= \frac{ -s_1-2w_{12}^2+s_2+2w_{22}^2 }{s_1} \\
 &= -1 + \frac{s_2}{s_1} - \frac{\beta}{s_1},
\end{align*}

and hence

\begin{eqnarray*}
\underbrace{\left(\frac{-c}{s_1}\right) + \underbrace{ \left(\frac{s_2}{s_1}\right) \underbrace{\cos(2r_2)}_{[-1,1]}}_{\left[\dfrac{-s_2}{s_1},\dfrac{s_2}{s_1}\right]}}_{}\quad \nonumber \\
\left[-1-\frac{\beta}{s_1}, -1-\frac{\beta}{s_1}+\frac{2s_2}{s_1}\right].
\end{eqnarray*}

By this, we have a non-empty solution set if and only if

\begin{equation}
(-1-\frac{\beta}{s_1} >= -1) \text{ and } (-1-\frac{\beta}{s_1}+\frac{2s_2}{s_1} <= 1).
\end{equation}

For the first predicate we require

\begin{align*}
-1 - \frac{2w_{12}^2-2w_{22}^2}{s_1} \leq 1, \\
-w_{12}^2+w_{22}^2 \leq s_1, \\
w_{22}^2 \leq w_{11}^2+w_{13}^2+w_{12}^2.
\end{align*}

The second predicate leads to a similar calculation and we obtain,
(harmlessly in our context) making the inequalities strict, here again
exactly the elegant condition (\ref{eqn:note49}). By this we see that
for the second and third cases of \(\msos{1}{1}\), if the set contained
minima it would also have to contain their related maxima, an altogether
`degenerate' situation compared to the minima coming from the curves
generated by \(\msos{2}{1}\). It is because of this that we spared
ourselves conducting the full second partial derivative test for that
case. Condition (\ref{eqn:note49}) is therefore a test for `correct' and
`useful' samples.

In fact there is a more direct explanation for the condition. It is easy
to see that \(p_i\) has its minimum at \(w_{i2}^2\) when \(\cos(r_i)\)
is zero and its maximum at \(\|w_i\|^2\) when the cosine is at an
extremum. At the same time, the error function \(d\) can only be zero
when there are \(\theta_i\) such that \(p_1=p_2\). But considering the
ranges above, this is equivalent to our condition. In other words,
samples that do not satisfy it are guaranteed to come from inexact
measurements. It goes without saying that the converse is not
necessarily true.

What the condition and its relation to \(\msos{1}{1}\) also tells us is
that the \(\msos{2}{1}\) must be a set of minima. Despite that, we
proceed to check this fact in a direct manner.

To do this, we feed the solution set into the relevant expressions by
doing the substitution

\begin{equation}
s_2\cos(2r_2) = c + s_1\cos(2r_1).
\end{equation}

\newcommand{\msico}[3]{ \left[\kern-0.5ex{#1}\atop{{\text{#2}}_{#3}}\kern-0.5ex\right] }
\newcommand{\msi}[2]{ \msico{#1}{si}{#2} }
\newcommand{\mco}[2]{ \msico{#1}{co}{#2} }

By this we obtain for \(\dfrac{\partial^2 O_1}{\partial \theta_1^2}\):

\begin{align*}
 & (-s_1) [2c \mco{2}{1} + 2s_1 \mco{4}{1} -2\mco{2}{1} (c + s_1\mco{2}{1} ) ] \\
=& (-s_1) [2s_1\mco{4}{1} -2s_1 (\frac{1}{2} + \frac{1}{2}\mco{4}{1} )] \\
=& (-s_1^2)(\mco{4}{1}-1).
\end{align*}

While for \(\dfrac{\partial^2 O_1}{\partial \theta_2^2}\) we have:

\begin{align*}
& 2c [s_2\mco{2}{2}] - 2s_2[s_2\mco{4}{2}] + 2s_1\mco{2}{1} [s_2\mco{2}{2}] \\
=& \underbrace{2c^2 + 4c s_1\mco{2}{1}}_{A} + \underbrace{2{s_1}^2 \mco{2}{1}^2}_{B} - \underbrace{2s_2 [s_2 \mco{4}{2}]}_{C}.
\end{align*}

With

\begin{align*}
B &= 2{s_1}^2(\frac{1}{2}+\frac{1}{2}\mco{4}{1}) \\
  &= {s_1}^2 + {s_1}^2\mco{4}{1},
\end{align*}

and

\begin{align*}
C &= 2(2{s_2} \mco{2}{2} {s_2} \mco{2}{2} - {s_2}^2) \\
  &= 2(2 [c+s_1 \mco{2}{1}]^2 - {s_2}^2) \\
  &= 4c^2 + 2B + 8c s_1 \mco{2}{1} - 2 {s_2}^2 \\
  &= 4c^2 + 2{s_1}^2+2{s_1}^2 \mco{4}{1}+8 c s_1 \mco{2}{1} - 2 {s_2}^2.
\end{align*}

We then reach for \(\dfrac{\partial^2 O_1}{\partial \theta_2^2}\) the
expression:

\begin{equation*}
 (-2c^2-s_1^2+2s_2^2) - (-s_1) (4c\mco{2}{1}+s_1\mco{4}{1}),
\end{equation*}

and consequently that

\begin{align*}
 \mdhex{L}{\msos{2}{1}} =& [(-s_1^2)(\mco{4}{1}-1)] \\ &\quad[(-2c^2-s_1^2+2s_2^2) - (-s_1) (4c\mco{2}{1}+s_1\mco{4}{1})].
\end{align*}

For \(\mdhex{R}{\msos{2}{1}}\) on the other hand, we obtain while
skipping a number steps the expression

\begin{align*}
 & 4(s_1\msi{2}{1})^2(s_2\msi{2}{2})^2 \\
 =& 2 (s_1\msi{2}{1})^2 (2s_2^2-s_2^2-s_2^2\mco{4}{2}) \\
 =& s_1^2(1-\mco{4}{1}) s_2^2 (1-\mco{4}{2}) \\
 =& s_1^2[1-{\kern-0.5ex}\mco{4}{1}{\kern-0.2ex}] [(-2c^2-{\kern-0.5ex}s_1^2+{\kern-0.5ex}2s_2^2) -{\kern-0.5ex}s_1(4c{\kern-0.5ex}\mco{2}{1}{\kern-0.8ex}-s_1{\kern-0.7ex}\mco{4}{1})].
\end{align*}

A swift comparison show that in fact this is the negative of
\(\mdhex{R}{\msos{2}{1}}\) and therefore

\begin{equation}
\mdhex{}{\msos{2}{1}} = 0.
\end{equation}

The test being inconclusive for our second stationary point set, we seek
a different path. Since this is a problem with a non-negative objective
functions, let us check if there is a relation between the zeros of the
objective function and the set at hand. The zeros of \(O_1\) occur at

\begin{align}
\left( [s_1\mco{1}{1}^2 + w_{12}^2] - [s_2\mco{1}{2}^2 + w_{22}^2] \right)^2 = 0, \nonumber \\
s_1(\frac{1}{2}+\frac{1}{2}\mco{2}{1} + w_{12}^2 = s_2(\frac{1}{2}+\frac{1}{2}\mco{2}{2} + w_{22}^2, \nonumber \\
\frac{s_1}{2}\mco{2}{1} - \frac{s_2}{2}\mco{2}{2} = -\frac{s_1}{2}-w_{12}^2+\frac{s_2}{2}+w_{22}^2.
\end{align}

Multiplying the above by two gives exactly \(\msos{2}{1}\) and this
confirms that this is a set of minima despite the inconclusive test.

Having done this, we proceed to a short study of the solution set's
curve. Since the cosine function is even, we notice that the curve must
be symmetric about the origin of the \((2r_1,2r_2)\) plane as well as
self-repeating with a period of \(2\pi\). Therefore to study it we only
need to focus on the \([0,\pi] \times [0,\pi]\) patch of the
aforementioned plane. Within this patch, when a solution set exists, it
is characterized by the monotone nature of \(\arccos\) since we then
have

\begin{equation}
2r_1 = \arccos\left(\frac{-c}{s_1} + \frac{s_2}{s_1}\cos(2r_2)\right).
\end{equation}

A canonical form of (\ref{eq:np55}) is

\begin{equation}
\cos{x} = u\cos(y)+v,\quad u > 0. \label{eq:s2canon}
\end{equation}

Since the function is monotone it is sufficient for a qualitative but
still exhaustive understanding of it to study its intersections with the
lines that delimit the patch. They are the lines \(x=0\), \(x=\pi\),
\(y=0\) and \(y=\pi\).

For \(x=0\) we have an intersection if and only if \([1 = u\cos(y)+v]\)
has solutions which leads to the condition
\[\frac{\lvert{1-v}\rvert}{u} \leq 1.\] A similar calculation gives the
following set of four individual conditions:

\begin{align*}
\frac{\lvert{1-v}\rvert}{u} &\leq 1, \\
\frac{\lvert{-1-v}\rvert}{u} &\leq 1, \\
\lvert{u+v}\rvert &\leq 1, \\
\lvert{-u+v}\rvert &\leq 1. \\
\end{align*}

\begin{figure}
\centering
\includegraphics[scale=0.44,trim={1.1cm 0.3cm 0.7cm 0},clip]{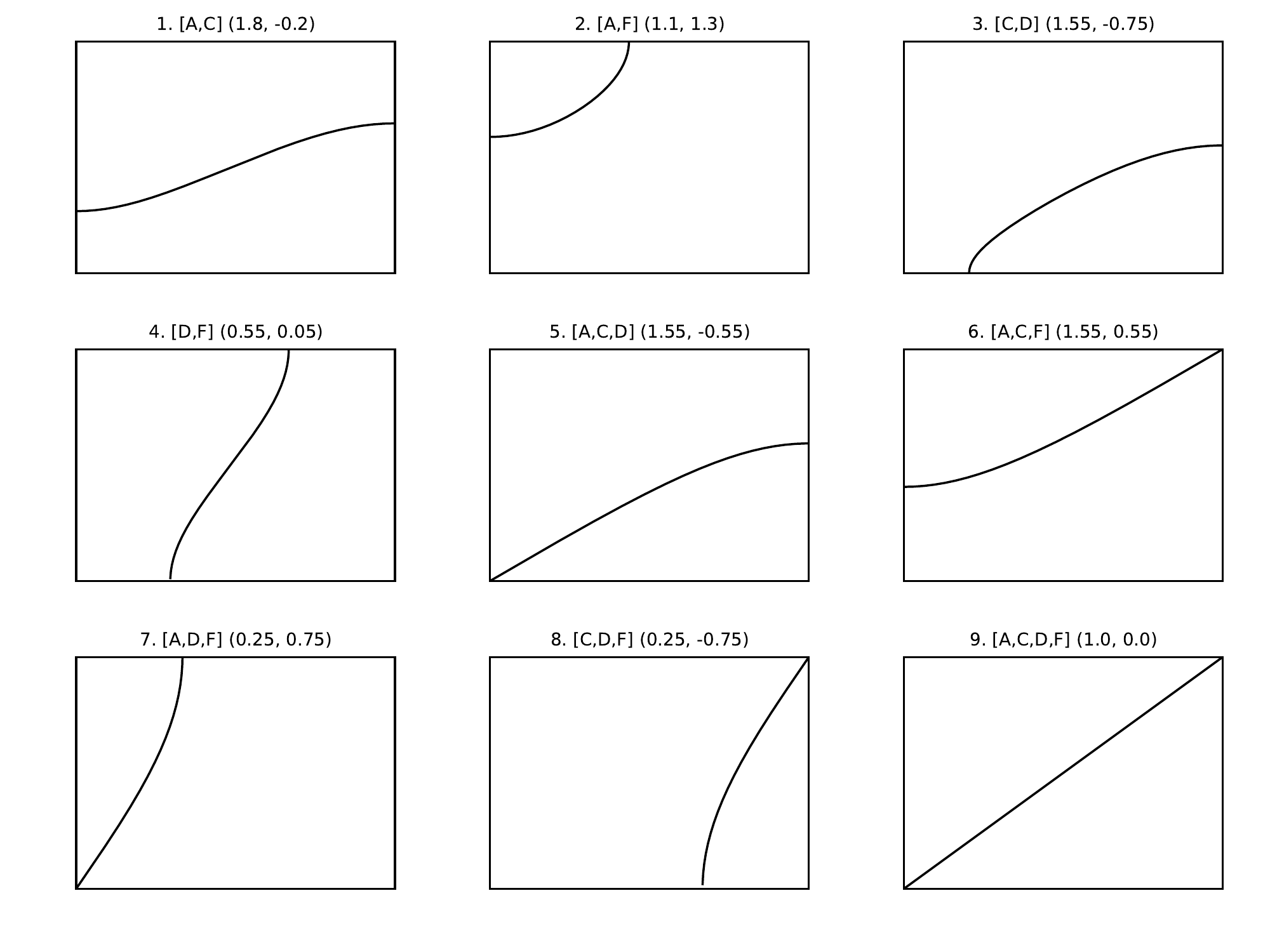}
\caption{The family of possible curves of $\mso{1}{2}$ within the patch $[0,\pi]\times[0,\pi]$ .}
\label{fig:script1}
\end{figure}

It is not very difficult to see that these conditions can be independent
but finding out which combinations of intersections are possible points
towards a dull path. We created a computer script (using Python) to do
the work. With the above conditions labelled as A,C,D,F, the results of
the script are that 9 combinations are possible out of the total 16.
This makes sense since this is the effect of the fact that \(b\) is
positive, allowing only half of the combinations, ignoring the last
`degenerate' one. The possible combinations along with witness
parameters for \(b\) and \(c\) and witness curves are provided in figure
(\ref{fig:script1})

Finally note that a calculus analysis can provide the same result by
showing that the slope of the curve within the patch is always positive.

\subsection{Uniqueness of Solutions}\label{uniqueness-of-solutions}

Having obtained decent understanding of the single-sample problem and
its characterizing solution curves we note that, in the absence of
noise, insight into multi-sample problem can be gained by considering
superpositions of the curves in the \((\theta_1,\theta_2)\) plane for
each of the samples. We have seen that each \(\msos{2}{1}\) is symmetric
about one of the \(\msos{1}{1}\) points. Due to this it becomes relevant
to pose the question of existence of samples that have unequal data,
equal \(\msos{1}{1}\) grids, but different \(\msos{2}{1}\) curves. When
such samples intersect within one of the four sectors of a patch around
a maximum, they necessarily intersect within all others creating three
false minima for each true one. If additionally the intersection point
can be quite arbitrary, the distances between the so obtained minima
would be arbitrarily small and this would make the problem ill-posed. We
proceed to find such two samples.

\newcommand{\msampled}[3]{ \tensor[^{\scriptscriptstyle{#3}}]{#1}{_{#2}} }
\newcommand{\mri}[2]{ \,\msampled{r}{#1}{#2} }
\newcommand{\mfracw}[2]{\frac{\tensor[^{\scriptscriptstyle{#2}}]{w}{_{#1{1}}}}{\tensor[^{\scriptscriptstyle{#2}}]{w}{_{#1{3}}}} }
\newcommand{\mvecw}[2]{\left(\msampled{w}{#1{1}}{#2}\atop\msampled{w}{#1{3}}{#2}\right)}
\newcommand{\mat}[2]{\text{atan}[\mfracw{#1}{#2}] }

For any two samples \(A\) and \(B\), looking first at \(\msos{1}{N}\),
we need \(\sin(2\mri{i}{A}) = 0\) if and only if
\(\sin(2\mri{i}{B}) = 0\), which reduces to

\begin{equation}
\mat{i}{A} = \mat{i}{B},\label{eqn:taneq}
\end{equation}

Since we write \(\text{atan}\) when we really mean \(\text{atan2}\),
this is equivalent to

\begin{equation}
\mvecw{i}{A} = \lambda_i\mvecw{i}{B} + 2k_i\pi,\quad \lambda_i > 0. \label{eqn:lbdaeq}
\end{equation}

This is then the condition for two samples to have identical
\(\msos{1}{2}\) grids. Having obtained this, we turn to \(\msos{2}{1}\)
looking for a family of curves passing through a specific point which we
force to lie on the \(x=y\) line to simplify finding an explicit
example. Using the canonical form (\ref{eq:s2canon}) we have at the
common point \((t,t)\) that

\begin{align*}
\cos(t) &= u\cos(t)+v, \\
\cos(t) &= \frac{v}{1-u}.
\end{align*}

Since the point is common to all curves its cosine must be equal in all
of them so we have for two samples \(A\) and \(B\) that

\begin{align}
\frac{\msampled{v}{}{A}}{1-\msampled{u}{}{A}} &= \frac{\msampled{v}{}{B}}{1-\msampled{u}{}{B}}. \nonumber \\
\msampled{u}{}{B} &= 1-\msampled{v}{}{B} \left[ \frac{1+\msampled{u}{}{A}}{\msampled{v}{}{A}} \right] \label{eqn:note122pp}.
\end{align}

Going back to the original form (\ref{eq:np55}) and baking in the
condition (\ref{eqn:lbdaeq}) (while dropping the \(2\pi\) period since
we are working within one patch) we have the two relations:

\begin{align*}
\msampled{v}{}{A} &= -1 + \left(\frac{{\lambda_2}^2}{{\lambda_1}^2}\right)\frac{\msampled{s}{2}{B}}{\msampled{s}{1}{B}} - \frac{\msampled{\beta}{}{A}}{({\lambda_1}^2)\msampled{s}{1}{B}},\\
\msampled{u}{}{A} &= \left(\frac{{\lambda_2}^2}{{\lambda_1}^2}\right) \frac{\msampled{s}{2}{B}}{\msampled{s}{1}{B}}.
\end{align*}

By (\ref{eqn:note122pp}) we then require the follwing relation between
the samples:

\begin{equation}
\frac{\msampled{s}{2}{B}}{\msampled{s}{1}{B}} = 1-
\left[\frac
{-1 + \frac{\msampled{s}{2}{B}}{\msampled{s}{1}{B}} - \frac{\msampled{\beta}{}{B}}{{\msampled{s}{1}{B}}}}
{-1 + \left(\frac{{\lambda_2}^2}{{\lambda_1}^2}\right)\frac{\msampled{s}{2}{B}}{\msampled{s}{1}{B}} - \frac{\msampled{\beta}{}{A}}{({\lambda_1}^2)\msampled{s}{1}{B}}}
\right]
\left[
1 + \left(\frac{{\lambda_2}^2}{{\lambda_1}^2}\right) \frac{\msampled{s}{2}{B}}{\msampled{s}{1}{B}}
\right] \label{eqn:note123}
\end{equation}

We denote

\begin{align*}
Q &= \frac{\msampled{s}{2}{B}}{\msampled{s}{1}{B}}, \\
R &= \frac{{\lambda_2}^2}{{\lambda_1}^2},
\end{align*}

and tentatively set

\begin{equation*}
R = \frac{1}{Q}
\end{equation*}

which simplifies (\ref{eqn:note123}) to

\begin{equation*}
R = 1-2\left[R-1-\frac{\msampled{\beta}{}{B}}{\msampled{s}{1}{B}}\right] \left[\frac{{(\lambda_1}^2)\msampled{s}{1}{B}}{-\msampled{\beta}{}{A}}\right]
\end{equation*}

Again tentatively setting

\begin{equation*}
R = 2
\end{equation*}

we obtain

\begin{align*}
\frac{{(\lambda_1}^2)\msampled{s}{1}{B}}{-\msampled{\beta}{}{A}} &= 2\left[1-\frac{\msampled{\beta}{}{B}}{\msampled{s}{1}{B}}\right], \\
\msampled{\beta}{}{B} &= \msampled{s}{1}{B} - \frac{1}{2{\lambda_1}^2}\msampled{\beta}{}{A}.
\end{align*}

To finally obtain our example we set

\begin{align*}
\msampled{s}{1}{B} &= 4,\\
{\lambda_1}^2 &= 4,\\
\msampled{\beta}{}{A} &= 4,\\
{\lambda_2}^2 &= 2,\\
\end{align*}

This determines the two samples almost completely and by inspection we
set

\begin{equation*}
\msampled{w}{22}{A} = \msampled{w}{22}{B} = 0,
\end{equation*}

to make the samples valid viz. satisfy (\ref{eqn:note49}). The samples
are then

\begin{align*}
\msampled{w}{1}{A},\msampled{w}{2}{A} &= (\sqrt{12},\sqrt{2},2),\,(\sqrt{2},0,\sqrt{14}), \\
\msampled{w}{1}{B},\msampled{w}{2}{B} &= (\sqrt{3},\sqrt{\frac{7}{4}},1),\,(1,0,\sqrt{7}), \\
\end{align*}

\begin{figure}
\centering
\includegraphics[scale=0.34,trim={1.1cm 0.3cm 0.7cm 0},clip]{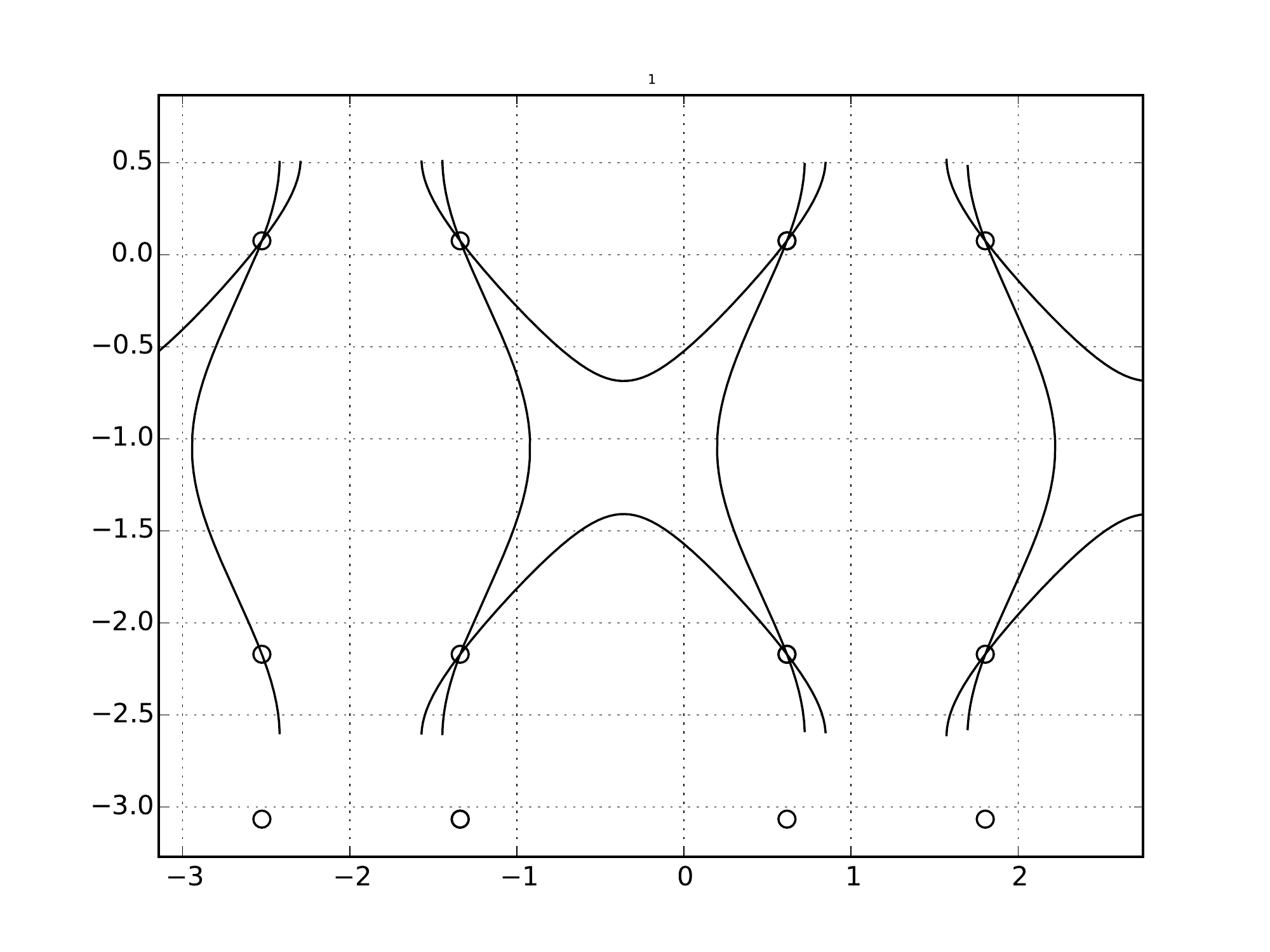}
\caption{Two $\msos{2}{1}$ curves intersecting at an arbitrarily chosen point. The minimum distance between the intersection points is less than $\frac{\pi}{3}$.The minima generated by running a non-linear least squares algorithm are shown using circles.}
\label{fig:script2}
\end{figure}

and the ensuing skinned cat figure (\ref{fig:script2}) confirms the
result.

Despite the barbarism of the attempt, we easily found our example. This
indicates that a continuum of such curves almost certainly exists but we
leave a proof of this as subject for further work. Thence, we have shown
that the problem is from the point of view of unique solutions,
ill-posed or more exactly, only conditionally well-posed.

\section{Appendix}\label{appendix}

\subsection{Objective Function
Intuition}\label{objective-function-intuition}

Very few sentences are used in (Seel, Schauer, and Raisch 2012) to
explain the derivation of the objective function. We here attempt an
explanation with the following intuitive perspective.

Consider a canonical hinge setup. It naturally leads to conceptually
thinking of the whole three-dimensional space as the union of two. The
`hinge plane' that is orthogonal to the hinge axis and in which the
constrained bodies canonically lie, and it's complement. The part of the
angular velocities that determine the planar rotation within the hinge
plane is given by projecting them onto the hinge axis. This exhausts the
hinge's single degree of freedom. The projection rests (also called
rejections) must therefore be equal. Another way to think about this is
that any difference in the rejections would `break' the plane for the
two canonical bodies would rotate such that they would not anymore lie
in a common plane. By known vector projection and rejection formulas,
with \(a\) being the unit hinge axis, \(\theta\) the hinge angle and
\(w_i\) being the angular velocities, we have:

\newcommand{\mwp}[1]{ \tensor[_ {\parallel}]{w}{_{#1}} }
\newcommand{\mwr}[1]{ \tensor[_ {\perp}]{w}{_{#1}} }

\begin{equation*}
\begin{split}
w_i = \underbrace{(w_i \cdot a) a}_{\mwp{i}} + \underbrace{a \times (w_i \times a)}_{\mwr{i}}. \\
\mwp{2} - \mwp{1} = \dot{\theta}a. \\
\mwr{2} - \mwr{1} = 0.
\end{split}
\end{equation*}

All of this holding in any space, our objective function simply
expresses the equality of rejections in two unknown spaces. Since the
spaces are unknown, the magnitude of both vectors is taken and the
spaces that make the equality hold determine the sought for
transformations.

\subsection{\texorpdfstring{Expressions Generated by
\emph{SymPy}}{Expressions Generated by SymPy}}\label{expressions-generated-by-sympy}

\begin{equation*}
\begin{split}
\frac{\partial O_1}{\partial \theta_1} &= (2 (w_{11} \sin(\theta_1)-w_{13} \cos(\theta_1)) (2 w_{11} \cos(\theta_1)\\
 &\quad +2 w_{13} \sin(\theta_1)) (w_{12}^2-w_{22}^2+(w_{11} \sin(\theta_1)\\
 &\quad -w_{13} \cos(\theta_1))^2-(w_{21} \sin(\theta_2)-w_{23} \cos(th2))^2)) \\
\frac{\partial^2 O_1}{\partial \theta_1^2} &=(2 (-2 w_{11} \sin(\theta_1)+2 w_{13} \cos(\theta_1)) (w_{11} \sin(\theta_1)\\
 &\quad -w_{13} \cos(\theta_1)) (w_{12}^2-w_{22}^2+(w_{11} \sin(\theta_1)\\
 &\quad -w_{13} \cos(\theta_1))^2-(w_{21} \sin(th2)-w_{23} \cos(th2)\\
 &\quad )^2)+2 (w_{11} \sin(\theta_1)-w_{13} \cos(\theta_1))^2 (2 w_{11} \cos(\theta_1)\\
 &\quad +2 w_{13} \sin(\theta_1))^2+2 (w_{11} \cos(\theta_1)+w_{13} \sin(\theta_1)\\
 &\quad ) (2 w_{11} \cos(\theta_1)+2 w_{13} \sin(\theta_1)) (w_{12}^2-w_{22}^2+(w_{11} \sin(\theta_1)\\
 &\quad -w_{13} \cos(\theta_1))^2-(w_{21} \sin(th2)-w_{23} \cos(th2))^2))
 \\
|\text{Hess }{O_1}| &= (-4 (w_{11} \sin(\theta_1)-w_{13} \cos(\theta_1))^2 (2 w_{11} \cos(\theta_1)\\
 &\quad +2 w_{13} \sin(\theta_1))^2 (w_{21} \sin(\theta_2)-w_{23} \cos(\theta_2)\\
 &\quad )^2 (2 w_{21} \cos(\theta_2)+2 w_{23} \sin(\theta_2))^2+(2 (-2 w_{11} \sin(\theta_1)\\
 &\quad +2 w_{13} \cos(\theta_1)) (w_{11} \sin(\theta_1)-w_{13} \cos(\theta_1)\\
 &\quad ) (w_{12}^2-w_{22}^2+(w_{11} \sin(\theta_1)-w_{13} \cos(\theta_1)\\
 &\quad )^2-(w_{21} \sin(\theta_2)-w_{23} \cos(\theta_2))^2)+2 (w_{11} \sin(\theta_1)\\
 &\quad -w_{13} \cos(\theta_1))^2 (2 w_{11} \cos(\theta_1)+2 w_{13} \sin(\theta_1)\\
 &\quad )^2+2 (w_{11} \cos(\theta_1)+w_{13} \sin(\theta_1)) (2 w_{11} \cos(\theta_1)\\
 &\quad +2 w_{13} \sin(\theta_1)) (w_{12}^2-w_{22}^2+(w_{11} \sin(\theta_1)\\
 &\quad -w_{13} \cos(\theta_1))^2-(w_{21} \sin(\theta_2)-w_{23} \cos(\theta_2)\\
 &\quad )^2)) (-2 (-2 w_{21} \sin(\theta_2)+2 w_{23} \cos(\theta_2)\\
 &\quad ) (w_{21} \sin(\theta_2)-w_{23} \cos(\theta_2)) (w_{12}^2-w_{22}^2+(w_{11} \sin(\theta_1)\\
 &\quad -w_{13} \cos(\theta_1))^2-(w_{21} \sin(\theta_2)-w_{23} \cos(\theta_2)\\
 &\quad )^2)+2 (w_{21} \sin(\theta_2)-w_{23} \cos(\theta_2))^2 (2 w_{21} \cos(\theta_2)\\
 &\quad +2 w_{23} \sin(\theta_2))^2-2 (w_{21} \cos(\theta_2)+w_{23} \sin(\theta_2)\\
 &\quad ) (2 w_{21} \cos(\theta_2)+2 w_{23} \sin(\theta_2)) (w_{12}^2-w_{22}^2+(w_{11} \sin(\theta_1)\\
 &\quad -w_{13} \cos(\theta_1))^2-(w_{21} \sin(\theta_2)-w_{23} \cos(\theta_2))^2)))
\end{split}
\end{equation*}

\section*{References}\label{references}
\addcontentsline{toc}{section}{References}

Seel, Thomas, Thomas Schauer, and J{ö}rg Raisch. 2012. ``Joint Axis and
Position Estimation from Inertial Measurement Data by Exploiting
Kinematic Constraints.'' In \emph{Control Applications (CCA), 2012 IEEE
International Conference on}, 45--49. IEEE.

SymPy Development Team. 2014. \emph{SymPy: Python Library for Symbolic
Mathematics}. \url{http://www.sympy.org}.

\end{document}